\documentclass{article}
\usepackage{spconf,amsmath,graphicx,subfigure}
\usepackage{booktabs}

\usepackage{multirow}
\usepackage{array}
\usepackage{xcolor}
\usepackage{chngpage}
\usepackage[symbol]{footmisc}

\title{Variants of BERT, Random Forests and SVM approach for Multimodal Emotion-Target Sub-challenge}

\name{Hoang Manh Hung,  
 Hyung-Jeong Yang, Soo-Hyung Kim, and Guee-Sang Lee$^{*}$\thanks{*Corresponding author}
}
\address{Department of Electronics and Computer Engineering, Chonnam National University, South Korea\\ 
}
\begin{document}
\maketitle
\begin{abstract}

Emotion recognition has become a major problem in computer vision in recent years that made a lot of effort by researchers to overcome the difficulties in this task. In the field of affective computing, emotion recognition has a wide range of applications, such as healthcare, robotics, human-computer interaction. Due to its practical importance for other tasks, many techniques and approaches have been investigated for different problems and various data sources. Nevertheless, comprehensive fusion of the audio-visual and language modalities to get the benefits from them is still a problem to solve. In this paper, we present and discuss our classification methodology for MuSe-Topic Sub-challenge, as well as the data and results. For the topic classification, we ensemble two language models which are ALBERT and RoBERTa to predict 10 classes of topics. Moreover, for the classification of valence and arousal, SVM and Random forests are employed in conjunction with feature selection to enhance the performance. 
\end{abstract}

\section{INTRODUCTION}
Emotion is an indispensable part of human experience. This has a major effect on human cognition, understanding, learning, communication, decision-making and many other scenarios of everyday life. One of the goals of the Affective Computing research is to make computers more human-like so that computers are aware of human emotional states. The emotion recognition has also been a great trend in psychology and in the computer science community because of its usefulness in practice \cite{C,alarcao2017emotions}. As an example, companies are interested in measuring people's comments about their services, and customers focus on feedback from other clients to evaluate the product before they purchase it. In the field of emotion recognition, there are many methods and modalities that have recently been used and focused on (i.e facial expression, speech, text, and physiological signals). The MuSe challenge \cite{stappen2020muse} aims to link the communities of Affective Computing and Sentiment Analysis and to compare the advantages of multimodal fusion for audio, vision, and language that assist emotion recognition systems to deal with behavior in the wild. The MuSe challenge includes 3 sub-challenges: Multimodal Sentiment in-the-Wild Sub-challenge (MuSe-Wild), Multimodal Emotion-Target Sub-challenge (MuSe-Topic), and Multimodal Trustworthiness Sub-challenge (MuSe-Trust). In this paper, we present our approach to the Muse-Topic sub-challenge, which is required to predict 10-class domain-specific topics as the emotional target and 3 levels of arousal and valence for each segment. 

\section{Methods}
\subsection{Topic classification}
In this task, 10 classes of the topics including (performance, interior-features, quality-aeshetic, comfort, handling, safety, general-information, cost, user-experience, exterior-features) will be used as labels to predict for each segment. From the experiments and also based on the provided baseline, we recognize that NLP transformer models outperform compared to other methods and modalities like Multimodal transformer (MMT) \cite{tsai2019multimodal}, LSTM \cite{hochreiter1997long}, End2You \cite{tzirakis2018end2you}. That is why we continue to exploit and use NLP transformer models. While BERT \cite{devlin2018bert} uses Next Sentence Prediction (NSP) and Masked Language Models (MLM) to help the network learn text representation, RoBERTa \cite{liu2019roberta} has shown that using NSP is inefficient. They used a dynamic masking pattern instead of a static one, replaced NSP by FULL-SENTENCES and DOC-SENTENCES approaches, trained with the bigger dataset and sentence length to improve performance. ALBERT \cite{lan2019albert} method tries to incorporate Factorized Embedding Parameterization and Cross Layer Parameter Sharing techniques to reduce the number of parameters to increase scalability for models. They also realized the inefficiency in the NSP and proposed a self-supervised loss for sentence-order prediction (SOP) to support the model focusing on inter-sentence coherence. For this task, we try to ensemble ALBERT and RoBERTa models trained with raw text from the transcripts to get the final result. 

\subsection{Emotion classification}
\begin{figure*}[ht]
\begin{center}
\includegraphics[scale=0.45]{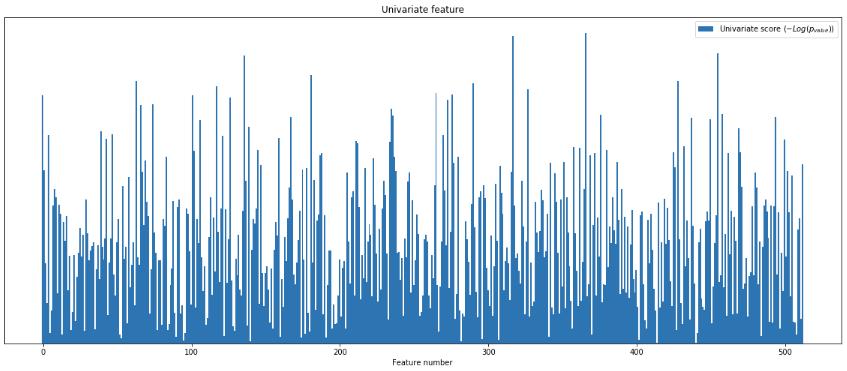}
\caption{The univariate values of each vggface feature. }
\label{fig:vgg}
\end{center}
\end{figure*}

\begin{figure*}[ht]
\begin{center}
\includegraphics[scale=0.45]{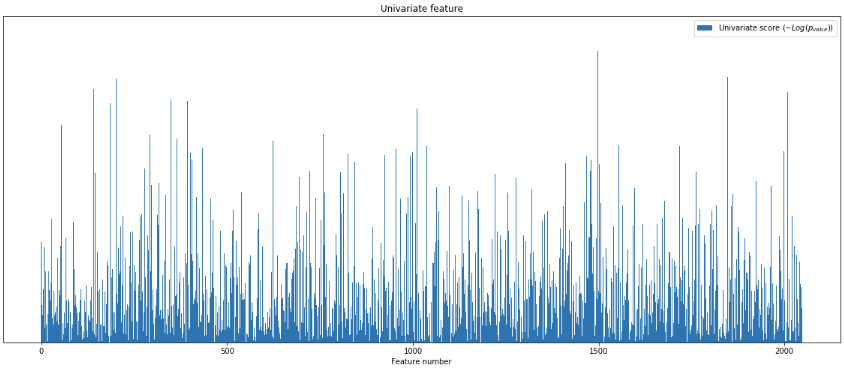}
\caption{The univariate values of each xception feature.}
\label{fig:xception}
\end{center}
\end{figure*}

Our approach is based on Support Vector Machines (SVMs) \cite{hearst1998support} and Random Forests to classify 3 levels of the emotion for both arousal and valence. The difference is that we use the univariate feature selection to select those features that have the strongest relationship with the output variable before the training (shown in Fig. \ref{fig:vgg} and Fig. \ref{fig:xception}). Based on the observation that there are numerical inputs and categorical outputs, we selected the ANOVA F-value method to perform statistical test. 


\section{Experimental results}
In the topic classification, ALBERT and RoBERTa use batch sizes of 16 and 32, respectively, with the sequence length are 50. The Adam optimizer is used along with the learning rate at the start of $10\sp{-5}$ and the total epochs are 3. The weights for the ensemble of two methods are respectively 0.5 and 0.5 for ALBERT and RoBERTa. We also try to use pseudo label for the training phase. The result is shown in the Table \ref{tab:result1}.

\begin{table}[ht]
\centering
\caption{Results of topic classification}
\label{tab:result1}
\begin{tabular}{@{}llll@{}}
\toprule
Feature                   & Method           & Valid         & Test           \\ \midrule
\multirow{5}{*}{Raw Text} & BaseLine(ALBERT) & 70.96         & 76.79          \\
                          & ALBERT           & 71.1          & \_             \\
                          & RoBERTa          & 73.7          & \_             \\
                          & ALBERT + RoBERTa & \textbf{74.5} & \textbf{77.65} \\
 & \begin{tabular}[c]{@{}l@{}}ALBERT + RoBERTa \\ + pseudo-labeling\end{tabular} & 73.8 & 77.07 \\ \bottomrule
\end{tabular}
\end{table}

For the emotion task, the complexity parameter C is set at 0.0538, and the gamma is automatic for SVM. The max depth of Random Forests method is set at 7.4008 and our best score is shown in Table \ref{tab:result2}.

\begin{table*}[t]
\caption{Results of emotion classification}
\centering
\label{tab:result2}
\begin{tabular}{@{}llllll@{}}
\toprule
                          &                         & \multicolumn{2}{l}{Arousal} & \multicolumn{2}{l}{Valence} \\ \midrule
Feature                   & Method                  & Valid         & Test        & Valid        & Test         \\ \midrule
\multirow{2}{*}{Deepspectrum} & KNeighborsClassifier             & 45.06          & \_              & \_             & \_             \\
                          & RandomForest            & \_            & \_          & 36.96        & \_           \\
\multirow{2}{*}{Egemaps}      & KNeighborsClassifier             & \_             & \_              & 36.34          & \_             \\
                          & RandomForest            & 44.21         & \_          & \_           & \_           \\
\multirow{2}{*}{Fasttext} & SVM                     & 45.48         & \_          & \_           & \_           \\
                          & RandomForest            & \_            & \_          & 38.05        & \_           \\
\multirow{2}{*}{Openpose} & RandomForest            & \_            & \_          & 37.81        & \_           \\
                          & RandomForest            & 46.82         & \_          & \_           & \_           \\
\multirow{2}{*}{Xception} & SVM                     & 46.61         & \_          & \_           & \_           \\
                              & RandomForest + feature selection & \_             & \_              & \textbf{39.82} & \textbf{39.75} \\
\multirow{2}{*}{Vggface}      & SVM + feature selection          & \textbf{50.35} & \textbf{45.164} & \_             & \_             \\
                          & SVM + feature selection & \_            & \_          & 38.12        & 31.91        \\ \bottomrule
\end{tabular}
\end{table*}

\section{Conclusion}
In this paper, we present and discuss our approach for MuSe-Topic sub challenge. We have shown the appropriate methods and parameters to produce high results in predicting topics and emotion states. However, there are still a lot of approaches that we have not experimented with, as well as exploited combinations of the modalities. They still have the potential to investigate for improving performance.

\bibliography{ref.bib}{}
\bibliographystyle{unsrt}
\end{document}